\documentclass[conference]{IEEEtran}
\IEEEoverridecommandlockouts

\usepackage{algorithm}
\usepackage{algpseudocode}
\usepackage{array}
\usepackage[caption=false,font=normalsize,labelfont=sf,textfont=sf]{subfig}
\usepackage{textcomp}
\usepackage{stfloats}
\usepackage{url}
\usepackage{verbatim}
\usepackage{graphicx}
\usepackage{cite}
\usepackage[hidelinks]{hyperref}

\usepackage{amsmath,amssymb,amsfonts, amsbsy, amsthm, bbm}
\usepackage{mathtools}
\usepackage{siunitx}
\usepackage{soul}
\usepackage{tabularx}
\usepackage{booktabs} %
\usepackage{adjustbox}
\usepackage[nameinlink,capitalize]{cleveref}
\crefname{figure}{Fig.}{Figs.}
\Crefname{figure}{Figure}{Figures}
\crefname{table}{Table}{Tables}
\Crefname{table}{Table}{Tables}
\crefname{section}{Section}{Sections}
\Crefname{subsection}{Section}{Sections}
\crefname{equation}{}{}      %
\Crefname{equation}{}{}
\creflabelformat{equation}{(#2#1#3)}
\crefrangelabelformat{equation}{(#3#1#4)--(#5#2#6)}

\renewcommand{\Pr}{\mathrm{P}}

\def\BibTeX{{\rm B\kern-.05em{\sc i\kern-.025em b}\kern-.08em
    T\kern-.1667em\lower.7ex\hbox{E}\kern-.125emX}}

\begin{document}

\title{Parallel Split Learning with Global Sampling}

\author{
  \IEEEauthorblockN{Mohammad Kohankhaki\textsuperscript{1}, 
  Ahmad Ayad\textsuperscript{2}, 
  Mahdi Barhoush\textsuperscript{3}, 
  and Anke Schmeink\textsuperscript{4}}\\
  \IEEEauthorblockA{
  Chair of Information Theory and Data Analytics\\
  RWTH Aachen University, D-52074 Aachen, Germany\\
   Email:\textsuperscript{1,}\textsuperscript{2,}\textsuperscript{3,}\textsuperscript{4}\{mohammad.kohankhaki, ahmad.ayad, mahdi.barhoush, anke.schmeink\}@inda.rwth-aachen.de}
}

\maketitle

\begin{abstract}
  Parallel split learning (PSL) suffers from two intertwined issues: the effective batch size grows with the number of clients, and data that is not identically and independently distributed (non-IID) skews global batches. We present parallel split learning with global sampling (GPSL), a server-driven scheme that fixes the global batch size while computing per-client batch-size schedules using pooled-level proportions. The actual samples are drawn locally without replacement by each selected client. This eliminates per-class rounding, decouples the effective batch from the client count, and makes each global batch distributionally equivalent to centralized uniform sampling without replacement. Consequently, we obtain finite-population deviation guarantees via Serfling’s inequality, yielding a zero rounding bias compared to local sampling schemes. GPSL is a drop-in replacement for PSL with negligible overhead and scales to large client populations. In extensive experiments on CIFAR-10/100 and ResNet-18/34 under non-IID splits, GPSL stabilizes optimization and achieves centralized-like accuracy, while fixed local batching trails by up to 60\%. Furthermore, GPSL shortens training time by avoiding inflation of training steps induced by data-depletion. These findings suggest GPSL is a promising and scalable approach for learning in resource-constrained environments.
\end{abstract}


\section{Introduction}\label{intro}

Distributed deep learning (DDL) has emerged as a powerful paradigm in the era of the Internet of Things (IoT), addressing the challenges posed by the proliferation of resource-constrained devices and the exponential growth of data at the network edge~\cite{parkWirelessNetworkIntelligence2019, zhouEdgeIntelligencePaving2019, parkDistillingOnDeviceIntelligence2019}. Split learning (SL) \cite{vepakommaSplitLearningHealth2018, poirotSplitLearningCollaborative2019} is a DDL paradigm designed for environments with constrained computational and communication resources, such as IoT settings. In SL, neural network training is partitioned between client devices and a central server. This setup enhances data privacy, minimizes client-side computation, and facilitates faster convergence \cite{vepakommaSplitLearningHealth2018, thapaAdvancementsFederatedLearning2021, liConvergenceAnalysisSequential2024}. However, the sequential execution across devices introduces considerable training and communication latency \cite{thapaAdvancementsFederatedLearning2021}.

To address training latency in SL, parallel split learning (PSL) \cite{jeonPrivacySensitiveParallelSplit2020, wuSplitLearningWireless2023, linEfficientParallelSplit2023, ohLocFedMixSLLocalizeFederate2022, hanAcceleratingFederatedLearning2021} has been proposed and introduces parallelism without the need for multiple server model instances like in split federated learning (v1) \cite{thapaSplitFedWhenFederated2022} or client model aggregation like in FedAvg \cite{mcmahanCommunicationEfficientLearningDeep2016}. However, PSL encounters the \textit{large effective batch size problem} \cite{palServerSideLocalGradient2021, ohLocFedMixSLLocalizeFederate2022}. As the client count grows, per-client batching with fixed local batch sizes makes the effective global batch size scale with the number of clients. This reduces gradient noise but can hurt generalization unless hyperparameters are carefully retuned \cite{golmantComputationalInefficiencyLarge2018,wilsonGeneralInefficiencyBatch2003}. It also increases server memory footprint and per-step latency.

Additionally, devices usually contribute from local datasets that are not independent and identically distributed (IID) and vary in sizes. This issue of \textit{non-IID data} is a general problem in DDL, not specific to PSL. Under non-IID data, local sampling schemes skew global batch composition due to rounding and per-client constraints, inflating deviation from the pooled distribution and slowing or destabilizing convergence. Rounding also accelerates \emph{client data depletion}, increasing the number of steps per epoch and prolonging training.

\begin{figure}[!t]
      \centering
      \includegraphics[width=0.78\columnwidth]{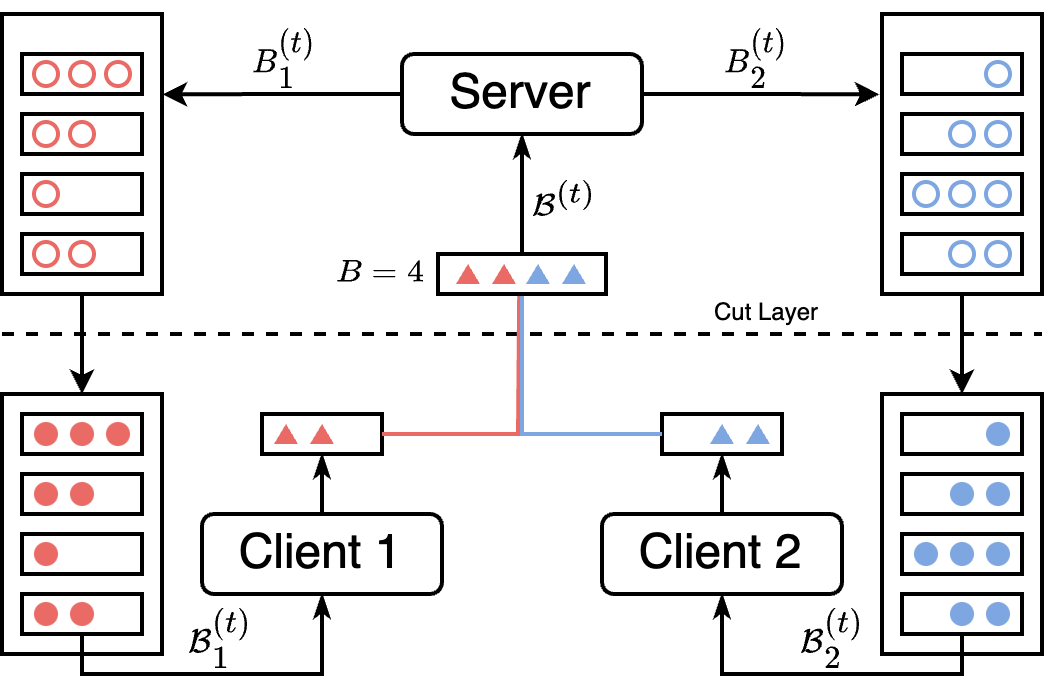}
  \caption{GPSL schematic: variable local batch sizes sum to a fixed global batch size, mitigating large effective batch size and non-IID issues in PSL.} 
      \label{fig_sim}
\end{figure}

We propose \emph{parallel split learning with global sampling} (GPSL) building upon the original PSL framework introduced in \cite{jeonPrivacySensitiveParallelSplit2020} to address these problems. It fixes a global batch size and assigns \emph{per-client batch-size schedules} derived from pooled-level proportions using only dataset-size metadata. Each client then samples its own data locally and without replacement to meet its scheduled count. This suppresses per-class rounding and decouples the effective batch size from the number of clients. Under GPSL, each global batch closely matches the pooled distribution in expectation and inherits finite-population deviation guarantees with zero rounding bias compared to local sampling schemes.

GPSL remains a drop-in replacement for PSL with negligible overhead. This means it integrates with existing PSL components such as client selection, clustering, and resource-aware schedulers \cite{liuWirelessDistributedLearning2023,linEfficientParallelSplit2023,palServerSideLocalGradient2021}. With a fixed global batch size and global sampling, it scales to large client populations without increasing server memory or per-step latency, and remains stable under non-IID splits. This makes it practical for edge learning: it requires only dataset-size metadata, uses lightweight client models with a single shared server model, and adds minimal coordination overhead.

An overview of GPSL is shown in \Cref{fig_sim}. The main contributions of this paper are:
\begin{itemize}
  \item \textbf{Novel sampling mechanism:} A server-driven \emph{global sampling} scheme that fixes the global batch size while allocating \emph{dynamic} local batch sizes, decoupling the effective batch size from the number of clients and eliminating per-class rounding effects present in local sampling schemes.
  \item \textbf{Finite-population guarantees:} We derive deviation bounds via Serfling’s inequality with the finite-population correction, yielding \emph{zero rounding bias} compared to local sampling schemes and showing that GPSL is equivalent to centralized uniform sampling without replacement.
\end{itemize}

The remainder of this paper is organized as follows. \Cref{related_work,preliminaries} present related works and preliminaries of our work. In \Cref{gpsl}, we introduce our novel GPSL method and analyze the deviation probability of global batches. \Cref{sec:simulation_results} presents the simulation results and addresses known limitations. \Cref{conclusion} concludes the paper with final remarks. Code and results available on GitHub~\cite{Mohkoh19GPSL}.

\section{Related Work}\label{related_work}

Jeon and Kim~\cite{jeonPrivacySensitiveParallelSplit2020} introduce PSL and use gradient averaging over all clients to address the client decoupling problem, where the information flow between clients is decoupled during back propagation (BP). They fixed the local batch sizes proportional to the dataset size of each client. Lyu et al.~\cite{lyuScalableAggregatedSplit2023} propose averaging the clients' local loss function outputs instead of their local gradients. This approach aligns with that of \cite{jeonPrivacySensitiveParallelSplit2020} due to the linearity of the gradient operator, but enhances privacy as clients can compute the local loss functions without sharing labels. Both show stable convergence and improved accuracy in both IID and non-IID settings. However, the proportion of each client's data in the total was chosen to yield an integer when scaled by the global batch size to determine local batch sizes. In practice, rounding is typically necessary when determining each client's local batch size, which can lead to imbalances in the global batch composition. Furthermore, the large effective batch size problem is not addressed in these works.

To mitigate the large effective batch size problem, Pal et al. \cite{palServerSideLocalGradient2021} propose scaling the learning rate on the server side while keeping the local batch size fixed and identical for all clients. Following the line of work from \cite{hanAcceleratingFederatedLearning2021}, Oh et al. \cite{ohLocFedMixSLLocalizeFederate2022} propose using smashed data augmentation, local loss function design and FedAvg for the client models to overcome the large effective batch size problem and client model decoupling. Pal et al. \cite{palServerSideLocalGradient2021} and Oh et al. \cite{ohLocFedMixSLLocalizeFederate2022} assume that the effective batch size scales with the number of clients. However, it has been shown that increasing the batch size leads to diminishing returns for gradient estimation \cite{golmantComputationalInefficiencyLarge2018} and that smaller batch sizes \cite{wilsonGeneralInefficiencyBatch2003} can be more beneficial for model generalization, if communication constraints allow. Furthermore, the empirical results are reported only for IID settings.

Cai and Wei \cite{caiEfficientSplitLearning2022a} propose an incentive mechanism to handle non-IID data in PSL, which encourages clients with larger and more evenly distributed datasets to participate in training. However, it introduces coordination and participation-scoring overhead and relies on stable client availability.
Other efforts to improve PSL focus on reducing training and communication latency \cite{wuSplitLearningWireless2023, linEfficientParallelSplit2023, zhangResourceefficientParallelSplit2024}. Methods include last-layer gradient aggregation, resource optimization, client clustering, and cut-layer selection.

\section{Preliminaries}\label{preliminaries}
Before introducing our GPSL method, we first present the system model and preliminaries of the PSL framework from \cite{jeonPrivacySensitiveParallelSplit2020} that we build upon. The client index set is $\mathcal{K}=\{1,2,\dots,K\}$. A server and $|\mathcal K|=K$ clients train a neural network model with parameter vector $\boldsymbol{w}$ collaboratively using the PSL framework. We focus on supervised classification tasks, where the dataset $\mathcal{D}_k$ of each client $k \in \mathcal{K}$ consists of input-label pairs $\mathcal{D}_k = \{(\boldsymbol{x}_{k,i}, y_{k,i})\}_{i=1}^{D_k}$, where $D_k \in \mathbb{N}$ is the dataset size, $\boldsymbol{x}_{k,i}$ is the input data, and $y_{k,i} \in \mathcal{M}$ is the ground truth class label of $\boldsymbol{x}_{k,i}$. Here, $\mathcal{M} = \{1, 2, \dots, M\}$ represents all classes and $M \in \mathbb{N}$ is the number of classes. The pooled dataset $\mathcal{D}_0 = \bigcup_{k \in \mathcal{K}} \mathcal{D}_k$ is the union of all clients' datasets, with a total size $D_0 = \sum_{k \in \mathcal{K}} D_k$. 

The \textit{class distribution} of a client $k$ with class label vector $\boldsymbol{y}_k \in \mathcal{M}^{D_k}$ is denoted as $\boldsymbol{\beta}_k \in \mathbb{R}^{M}$ with 

\begin{equation}
{\boldsymbol{\beta}_k = \frac{1}{D_k}(c(\boldsymbol{y}_k,1), c(\boldsymbol{y}_k,2), \dots, c(\boldsymbol{y}_k,M))^\top},
\label{eq:class_distribution}
\end{equation}

\noindent where $c(\boldsymbol{y}, m)$ counts the number of times class $m$ appears in $\boldsymbol{y}$. The \textit{pooled class distribution} of $\mathcal{D}_0$ with class label vector $\boldsymbol{y}_0 \in \mathcal{M}^{D_0}$ is denoted as $\boldsymbol{\beta}_0 \in \mathbb{R}^{M}$ according to the same definition.

Every epoch consists of $T \in \mathbb{N}$ optimization steps. The set of optimization steps is denoted as $\mathcal{T} = \{1, 2, \ldots, T\}$. Each client $k$ has a \emph{local batch size} $B_k^{(t)} \in \mathbb{N}$. The \emph{global batch size} $B^{(t)} = \sum_{k \in \mathcal{K}} B_k^{(t)}$ is the total number of intermediate representations that arrive at the server in each optimization step $t \in \mathcal{T}$. Accordingly, we differentiate between local batches $\mathcal{B}^{(t)}_k \subseteq \mathcal{D}_k$ and global batches $\mathcal{B}^{(t)} = \bigcup_{k \in \mathcal{K}} \mathcal{B}^{(t)}_k$.

PSL splits the model into client-side parameters $\boldsymbol{w}_c$ and server-side parameters $\boldsymbol{w}_s$. Before training, the server sends $\boldsymbol{w}_c$ to the clients, each storing a copy $\boldsymbol{w}_{c,k}$ with identical initial values. Accordingly, the model function $f(\boldsymbol{x}; \boldsymbol{w})$ is split into two functions $f_c$ and $f_s$ as $f(\boldsymbol{x}; \boldsymbol{w}) = f_s(f_c(\boldsymbol{x}; \boldsymbol{w}_c); \boldsymbol{w}_s)$. 
The goal is to find the optimal model parameters ${\boldsymbol{w}^* = (\boldsymbol{w}_s^*, \boldsymbol{w}_c^*)^\top}$ that minimize the global loss function. Each optimization step $t \in \mathcal{T}$ involves the following substeps:

\begin{enumerate}
  \item \textbf{Parallel Client-Side Forward Propagation (FP):} Each client $k$ samples a local batch $\mathcal{B}^{(t)}_k \subseteq \mathcal{D}_k$ with a local batch size of $B_k^{(t)}$ and sends $f_c(\mathcal{B}^{(t)}_k; \boldsymbol{w}_{c,k})$ with the corresponding labels to the server.
  \item \textbf{Server-Side FP:} The server concatenates the received intermediate representations to obtain the global batch ${\mathcal{B}^{(t)} = \bigcup_{k \in \mathcal{K}} f(\mathcal{B}^{(t)}_k; \boldsymbol{w}_{c,k})}$ and computes the final predictions $f_s(\mathcal{B}^{(t)}; \boldsymbol{w}_s)$ to compute the average loss.
  \item \textbf{Server-Side Backward Propagation (BP):} The server computes the server-gradients through backpropagation, updates its model parameters according to the optimization algorithm, and sends the gradient of the first server-side layer (\textit{cut layer}) to all clients.
  \item \textbf{Parallel Client-Side BP:} Each client $k$ continues backpropagation using the cut layer gradient to compute the client-gradients $\boldsymbol{g}^{(t)}_{\boldsymbol{w}_{c,k}}$ and sends it back to the server.
  \item \textbf{Server-Side Gradient Averaging:} The server computes the average client gradients $\boldsymbol{\overline{g}}^{(t)}_{\boldsymbol{w}_c} = \sum_{k \in \mathcal{K}} \frac{D_k}{D_0} \boldsymbol{g}^{(t)}_{\boldsymbol{w}_{c,k}}$ and sends it to all clients.
  \item \textbf{Parallel Client-Side Model Update:} Each client uses $\boldsymbol{\overline{g}}^{(t)}_{\boldsymbol{w}_c}$ to update its model parameters.
\end{enumerate}
An epoch is concluded after $T$ optimization steps. $T$ is determined by the number of non-empty global batches that can be formed from the clients' datasets until depletion and thus depends on the sampling procedure. The model is trained for a fixed number of epochs $E \in \mathbb{N}$. 

We adopt a generalized PSL formulation that permits variable local batch sizes across $k$ and $t$ to ensure compatibility with the proposed method. In standard PSL, by contrast, the local batch size is typically fixed over time, i.e., $B_k^{(t)} = B_k$ for all $t \in \mathcal{T}$ \cite{jeonPrivacySensitiveParallelSplit2020, lyuScalableAggregatedSplit2023, palServerSideLocalGradient2021}. Likewise, if the global batch size is fixed, we write $B^{(t)} = B$. When the per-client sequence $\{B_k^{(t)}\}_{t \in \mathcal{T}}$ varies with $t$, we refer to it as the per‑client batch‑size schedule or simply \emph{schedule}.

\section{System Model} \label{gpsl}
This section introduces the proposed GPSL method and analyzes how it improves the alignment between global batches and the pooled data distribution compared to random uniform sampling in a centralized setting.

\subsection{Parallel Split Learning with Global Sampling}
\label{global_uniform_sampling}

In GPSL, we replace the notion of fixed per‐client batch sizes $B_k$ with per-client schedules $\{B_k^{(t)}\}_{t \in \mathcal{T}}$ that sum across clients to a fixed \emph{global batch size} \(B\) in each step $t$. The server first gathers each client’s dataset size \(D_k\). This information could trivially be obtained from the server through counting the number of samples that arrive from each client and is not sensitive.  Before training the server must choose how many samples \(B_k^{(t)}\) each client \(k\) will contribute to each training step \(t \in \mathcal{T}\), subject to
\[
  \sum_{k \in \mathcal{K}} B_k^{(t)} \;=\; B.
\]
Rather than uniformly fixing these \(B_k^{(t)}\), the server computes local batch sizes by sampling \emph{client indices} using the pooled proportions implied by the remaining counts, i.e., the number of unused samples per-client. No raw data is accessed by the server and clients later draw their own examples locally to meet the assigned counts. We initialize the number of remaining unused datapoints per-client as \(R_k \gets D_k\) for all \(k \in \mathcal{K}\). For each step \(t \in \mathcal{T}\), the server allocates the per-step \emph{schedule} $\{B_k^{(t)}\}_{k\in\mathcal{K}}$ across clients according to the pooled proportions using only the remaining counts $\{R_k\}_{k\in\mathcal{K}}$: 
\\\\
\noindent For \(i=1,\dots,B\):
\begin{enumerate}
  \item Compute \emph{client weights}
  \begin{equation}
  \pi_k = \frac{R_k}{\sum_{j \in \mathcal{K}} R_j} .
  \label{eq:client_weights}
  \end{equation}
  \item Sample a client index \[z \sim \mathrm{Categorical}(\pi_1,\dots,\pi_K).\]
  \item Increment the scheduled count \(B_{z}^{(t)} \leftarrow B_{z}^{(t)} + 1\).
  \item Decrement the remaining count \(R_{z} \leftarrow R_{z} - 1\).
  \label{eq:decrement_dataset}
\end{enumerate}

\noindent 
The server precomputes and broadcasts the schedules \(\{B_k^{(t)}\}_{t \in \mathcal{T}}\) to all clients. At step \(t\), client \(k\) samples \(B_k^{(t)}\) previously unused examples uniformly without replacement from \(\mathcal{D}_k\) and then performs the regular PSL steps (see \cref{preliminaries}). A final partial batch (if any) may be kept or skipped. No other part of the PSL loop changes, so this sampling procedure drops into existing PSL implementations.

The full procedure (including handling dataset depletion across optimization steps) is summarized in \cref{alg:gpsl_sampling}. The time complexity is $O(BT)$, identical (up to constants) to fixed local batching.

\begin{algorithm}[t]
  \caption{Global Sampling Procedure for GPSL}
  \label{alg:gpsl_sampling}
  \begin{algorithmic}[1]
    \Require Number of clients $K$; client dataset sizes $D_1,\dots,D_K$; global batch size $B$.
  \Ensure Local schedules $\{ B_k^{(t)} \}_{t \in \mathcal{T}}$ for each \(k \in \mathcal{K}\).
    \State $R_k \gets D_k$ for all $k\in \mathcal{K}$ \Comment{Remaining samples}
    \State $t \gets 1$
    \While{$\sum_{k \in \mathcal{K}} R_k > 0$}
  \State $B_k^{(t)} \gets 0$ for all $k \in \mathcal{K}$ \Comment{Initialize scheduled counts}
        \State $L \gets \min\big(B, \sum_{j=1}^{K} R_j\big)$ \Comment{Number of draws this step}
        \For{$i = 1$ to $L$}
            \State $\pi_k \gets R_k / \sum_{j=1}^{K} R_j$ for all $k \in \mathcal{K}$ with $R_k>0$.
            \State $z \sim \mathrm{Categorical}(\pi_1,\dots,\pi_K)$.
            \State $B_{z}^{(t)} \gets B_{z}^{(t)} + 1$.
            \State $R_{z} \gets R_{z} - 1$.
        \EndFor
        \State $t \gets t + 1$
    \EndWhile
    \State Send $\{ B_k^{(t)} \}_{k \in \mathcal{K}}$ to all clients.
  \end{algorithmic}
\end{algorithm}

\subsection{Deviation Analysis}
\label{deviation_analysis}

We assess how well a batch \(\mathcal{B}\subset \mathcal{D}_0\) of size \(B\) reflects the true pooled class distribution \(\boldsymbol\beta_0\) using the $\ell_1$ deviation:
\begin{equation}
  \Delta(\mathcal{B},\boldsymbol\beta_0)
  = \sum_{m \in \mathcal{M}} \bigl|\hat\beta_m - \beta_{0,m}\bigr|,
  \quad
  \hat\beta_m = \frac{1}{B}\,c(\mathcal{B},m).
  \label{eq:deviation_function}
\end{equation}
where \(c(\mathcal{B},m)\) is the count of class-\(m\) samples in \(\mathcal{B}\). Here, \(\boldsymbol{\beta}_0\) denotes the pooled class distribution defined in \cref{preliminaries} and \(D_0 = \sum_{k\in\mathcal K} D_k\) is the total number of samples across all clients. Throughout we assume
\begin{equation}
  B \le D_0,
  \quad
  0 < \varepsilon \le 1 - \frac{1}{D_0},
  \label{eq:assumptions}
\end{equation}
so that: (i) the finite–population correction term \(1 - \tfrac{B-1}{D_0} > 0\) appearing in Serfling’s inequality remains positive (this requires \(B \le D_0\)); (ii) the deviation event \(\lvert Y_m - B\beta_{0,m}\rvert \ge \varepsilon B\) is non-vacuous because the maximal attainable deviation for a count bounded in \([0,B]\) is strictly less than \(B\) (hence we enforce \(\varepsilon B \le B-1\)); and (iii) choosing \(\varepsilon > 1 - 1/D_0\) would correspond to ruling on deviations that cannot occur under a hypergeometric draw (since at least one sample of any present class must exist in \(\mathcal{D}_0\)). These mild constraints avoid degenerate parameter settings and keep the bound informative. When \(\mathcal{B}\) is drawn by sampling \(B\) points without replacement from \(\mathcal{D}_0\), the count
\(\displaystyle Y_m = c(\mathcal{B},m)\)
follows a hypergeometric distribution 
\begin{equation}
  Y_m \sim \mathrm{Hypergeometric}(D_0,\,c(D_0,m),\,B).
  \label{eq:hypergeometric} \nonumber
\end{equation}
\noindent
with \mbox{$\mathbb{E}[Y_m]=B\,\beta_{0,m}$}. By Serfling’s inequality with finite‐population correction~\cite{Serfling1974},

\begin{align}
  \Pr\bigl(|Y_m - B\beta_{0,m}|\ge \varepsilon B\bigr)
  &= \Pr\!\Bigl(\bigl|\tfrac{Y_m}{B}-\beta_{0,m}\bigr|\ge \varepsilon\Bigr) \\
  &\le 2\exp\!\Bigl(-\frac{2\,\varepsilon^2\,B}{\,1-\tfrac{B-1}{D_0}\,}\Bigr).
\end{align}
Although the \(\{Y_m\}_{m \in \mathcal{M}}\) are dependent as they are required to sum up to \(B\), if \(\Delta(\mathcal B,\boldsymbol\beta_0)=\sum_{m\in\mathcal M}|Y_m/B-\beta_{0,m}|\ge M\varepsilon\), then for some \(m\) we have \(|Y_m/B-\beta_{0,m}|\ge\varepsilon\). A union bound over \(m\) yields
\begin{align}
  \Pr\!\Bigl(\Delta(\mathcal{B},\boldsymbol\beta_0)\ge M\varepsilon\Bigr)
  &\le \sum_{m \in \mathcal{M}} 2\exp\!\Bigl(-\tfrac{2\,\varepsilon^2\,B}{1-\tfrac{B-1}{D_0}}\Bigr)
  \\
  &= 2M\exp\!\Bigl(-\tfrac{2\,\varepsilon^2\,B}{1-\tfrac{B-1}{D_0}}\Bigr).
  \label{eq:union_bound}
\end{align}

In \cref{alg:gpsl_sampling}, at each draw the server samples a \emph{client index} \(k\) with probability \(\pi_k = R_k / \sum_{j\in\mathcal K} R_j\), conceptually decrements one remaining item for that client, and updates the counts \(\{R_k\}_{k\in\mathcal{K}}\). Conditional on the current \(\{R_k\}_{k\in\mathcal{K}}\), this index-sampling is \emph{analytically equivalent} to selecting one element uniformly at random from the pooled set of unused samples, even though the server never accesses raw examples. Clients later draw their own data to fulfill the assigned counts. Iterating this for \(B\) draws yields a uniform random sample of size \(B\) without replacement at the pooled level. Consequently, for any class \(m\), the batch count \(Y_m = c(\mathcal B,m)\) under GPSL is hypergeometric with the same parameters as centralized sampling, with \(\mathbb E[Y_m] = B\,\beta_{0,m}\). The joint vector \((Y_m)_{m\in\mathcal M}\) thus depends only on \((D_0, \boldsymbol{\beta}_0)\) and not on the client partition \(\{D_k\}_{k\in\mathcal{K}}\) or its skew. It follows that the finite-population Serfling bound in \Cref{eq:union_bound} applies to GPSL verbatim.

\subsection{Comparison with Fixed Local Batches}
\label{sec:comparison_fixed_local_batches}

Consider fixed local batch sizes \(\tilde B_k=\lceil B\,D_k/D_0\rceil\) sampled without replacement per client~\cite{jeonPrivacySensitiveParallelSplit2020}. Let \(\tilde Y'_m=\sum_{k \in \mathcal{K}} \tilde Y_{k,m}\) with
\[\tilde Y_{k,m}\sim\mathrm{Hypergeometric}(D_k,c(D_k,m),\tilde B_k)\]
and define
\[
  \tilde p_m \coloneqq \sum_{k \in \mathcal{K}}\frac{\tilde B_k}{B}\,\beta_{k,m}.
\]
The triangle inequality gives
\begin{equation}
  \bigl|\tfrac{\tilde Y'_m}{B}-\beta_{0,m}\bigr|
  \le
  \underbrace{\bigl|\tfrac{\tilde Y'_m}{B}-\tilde p_m\bigr|}_{\text{sampling noise}}
  +
  \underbrace{\bigl|\tilde p_m-\beta_{0,m}\bigr|}_{\text{rounding bias}}.
  \label{eq:decomposition}
\end{equation}
The sampling term obeys the same Serfling tail as above (conditioned on \(\{\tilde B_k\}_{k \in \mathcal{K}}\)).
For the deterministic rounding term, since \(0\le\beta_{k,m}\le 1\),
\begin{equation}
  \bigl|\tilde p_m-\beta_{0,m}\bigr|
  = \Bigl|\sum_{k \in \mathcal{K}}\Bigl(\tfrac{\tilde B_k}{B}-\tfrac{D_k}{D_0}\Bigr)\beta_{k,m}\Bigr|
  \le \sum_{k \in \mathcal{K}} \tfrac{1}{B}
  = \tfrac{K}{B}.
  \label{eq:delta_bound}
\end{equation}
Let \(\delta\coloneqq \max_{m \in \mathcal{M}} |\tilde p_m-\beta_{0,m}|\le\tfrac{K}{B}\). Combining \Cref{eq:decomposition}–\Cref{eq:delta_bound} and union bounding over \(m\) gives the \emph{shifted} tail bound
\begin{equation}
  \Pr\!\Bigl(\Delta(\mathcal{B},\boldsymbol\beta_0)\ge M\varepsilon\Bigr)
  \le
  2M\exp\!\Bigl(-\tfrac{2\,(\varepsilon-\delta)^2\,B}{\,1-\tfrac{B-1}{D_0}\,}\Bigr).
  \label{eq:shifted_tail_bound}
\end{equation}
Hence concentration controls only the random sampling fluctuations. Rounding introduces a fixed bias \(\delta\) that does not decay exponentially and can dominate when \(K/B\) is not negligible. GPSL avoids this bias (\(\delta=0\)) while retaining the same exponential decay as \Cref{eq:union_bound}. Fixed local batches may still perform adequately when \mbox{\(K \ll B\)} and rounding effects are negligible. However, such settings are rare in edge learning environments, where large \(K\) and small batch sizes dominate due to memory and communication constraints. Moreover, the rounding exacerbates the issue of \emph{client data depletion}, where client datasets are depleted disproportionately and faster. This increases the total number of training steps \(T\) and prolongs training. By contrast, GPSL prevents unnecessary inflation of \(T\) by keeping the global batch size \(B\) constant, regardless of the number of clients.

\section{Simulation Results}
\label{sec:simulation_results}

In this section, we empirically evaluate the performance of GPSL compared to existing sampling methods in PSL. We examine the impact of sampling on batch deviation, training latency, and classification accuracy under both IID and non-IID data splits.

\subsection{Experimental Setup}
We use the CIFAR-10 dataset \cite{krizhevskyLearningMultipleLayers2009}, which consists of $60,000$ color images in $M=10$ classes. The dataset is split into $D_0 = 50,000$ training and $10,000$ test images.

We train a \emph{ResNet-18} \cite{heDeepResidualLearning2016} for $E=100$ epochs with mini-batch SGD and momentum. The cut layer is the third residual block on the server. To accommodate variable local batches, client-side models use GroupNorm \cite{wuGroupNormalization2020} with 32 groups, while the server-side model retains BatchNorm \cite{ioffeBatchNormalizationAccelerating2015}. Unless otherwise stated, we use a \emph{fixed global batch size} $B=128$ and $K=64$ clients in all experiments; other settings are varied explicitly. We run each configuration 5 times with different seeds and report mean$\,\pm$\,standard deviation. Details about the experimental setup are summarized in \cref{tab:experimental_setup}. Additional experiments on the CIFAR-100 dataset \cite{krizhevskyLearningMultipleLayers2009} and ResNet-34 are provided in \Cref{app:additional_results}.

\begin{table}[t]
\caption{Experimental setup.}
\label{tab:experimental_setup}
\centering
\begin{tabularx}{\columnwidth}{l>{\raggedright\arraybackslash}X}
\toprule
\textbf{Aspect} & \textbf{Configuration} \\
\midrule
Dataset & CIFAR-10: 50k train / 10k test, 10 classes \\
Splits & IID; Non-IID (mild: $C{=}5,\,\alpha{=}3.0$), (severe: $C{=}2,\,\alpha{=}3.0$) \\
Clients & $K\in\{16,32,64,128\}$ (default $K{=}64$) \\
Global batch & $B\in\{64,128,256\}$ (default $B{=}128$) \\
Cut layer & Residual layer 3 (server) \\
Optimizer & SGD with momentum $0.9$, weight decay $5\times10^{-4}$, learning rate $10^{-2}$ \\
Loss & Cross entropy with label smoothing ($\epsilon=0.1$) \\
Epochs & $E{=}100$ \\
Augmentation & Random crop, random horizontal flip \\
\bottomrule
\end{tabularx}
\end{table}

\subsection{Sampling Methods}
\label{sec:sampling_methods}
We compare GPSL against two sampling methods commonly used in PSL:

\begin{itemize}
    \item \textbf{Fixed local sampling (FLS)} \cite{palServerSideLocalGradient2021}: 
    Each client $k$ has the same local batch size $B_k = \lceil B / K \rceil$.
    \item \textbf{Fixed proportional sampling (FPLS)} \cite{jeonPrivacySensitiveParallelSplit2020}: 
    Each client $k$ has a proportional $B_k \propto D_k$ with $B_k = \lceil B \cdot \frac{D_k}{D_0} \rceil$.
\end{itemize}

We also include a \emph{centralized learning} (CL) baseline using random uniform sampling, where all data are pooled at the server and trained with a single model. This serves as an upper bound for performance. 

In \Cref{tab:iid_results} we report the results of the sampling methods in the IID setting, where all clients have the same distribution. All sampling methods perform similarly, achieving performance on par with the centralized learning (CL) baseline.

\subsection{Non-IID Data}
\label{sec:noniid_data}
\begin{figure}[!t]
      \centering
      \includegraphics[width=0.9\columnwidth]{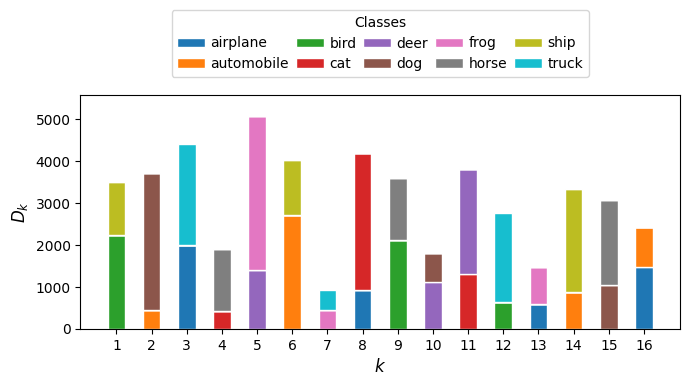}
      \caption{Example of a severe non-IID setting for $K=16$ clients, showing the distribution of classes across clients.}
      \label{fig:ex_dir_example}
\end{figure}

In the non-IID scenario, we employ a modified version of the extended Dirichlet allocation method proposed in~\cite{liConvergenceAnalysisSequential2024}, which ensures that each client receives data from exactly $C$ classes. First, $C$ distinct classes are assigned to each client using a multinomial-based allocation that promotes balanced class coverage across the client population. Then, for each class, a Dirichlet distribution with concentration parameter $\alpha$ is sampled to determine how its samples are distributed among the clients assigned to it. Lower values of $\alpha$ lead to more imbalanced allocations, where a class is concentrated heavily on one or a few clients. Higher values of $\alpha$ result in more uniform class sample distributions across clients. By tuning $C$ and $\alpha$, we can control the degree and structure of statistical heterogeneity across clients.
Thus, $C$ governs the per-client label diversity (a hard cap on how many distinct classes a client can observe), while $\alpha$ controls the smoothness of intra-class sharing across the eligible clients. The pair $(C,\alpha)$ therefore parameterizes the severity and structure of the data heterogeneity.
We consider a \emph{mild} non-IID scenario with $C=5$ and $\alpha=3.0$ and a \emph{severe} non-IID scenario with $C=2$ and $\alpha=3.0$. In~\Cref{fig:ex_dir_example} we show an example of the class distribution for $K=16$ clients in the severe non-IID scenario.

\renewcommand{\arraystretch}{1.2}
\begin{table}[t]
    \centering
    \caption{IID Test Accuracy}
    \label{tab:iid_results}
    \begin{tabularx}{\columnwidth}{l*{4}{>{\centering\arraybackslash}X}}
        \toprule
        \textbf{Method} & $\mathbf{K=16}$ & $\mathbf{K=32}$ & $\mathbf{K=64}$ & $\mathbf{K=128}$ \\
        \midrule
        FLS   & \mbox{84.21 $\pm$ 0.50} & \mbox{84.31 $\pm$ 0.54} & \mbox{84.15 $\pm$ 0.79} & \mbox{84.10 $\pm$ 0.36} \\
        FPLS  & \mbox{84.42 $\pm$ 0.58} & \mbox{84.31 $\pm$ 0.54} & \mbox{84.15 $\pm$ 0.79} & \mbox{84.10 $\pm$ 0.36} \\
        GPSL  & \mbox{\textbf{84.74} $\pm$ 0.47} & \mbox{\textbf{84.40} $\pm$ 0.56} & \mbox{\textbf{84.29} $\pm$ 0.53} & \mbox{\textbf{84.66} $\pm$ 0.45} \\
        \bottomrule
    \end{tabularx}

    \vspace{1.5em}

    \caption{Severe non-IID Test Accuracy $(C=2, \alpha=3.0)$}
    \label{tab:results_cifar10_severe}
    \begin{tabularx}{\columnwidth}{l*{4}{>{\centering\arraybackslash}X}}
        \toprule
        \textbf{Method} & $\mathbf{K=16}$ & $\mathbf{K=32}$ & $\mathbf{K=64}$ & $\mathbf{K=128}$ \\
        \midrule
        FLS   & \mbox{61.55 $\pm$ 7.18} & \mbox{57.17 $\pm$ 4.33} & \mbox{66.02 $\pm$ 6.00} & \mbox{67.73 $\pm$ 9.29} \\
        FPLS  & \mbox{59.09 $\pm$ 8.25} & \mbox{64.88 $\pm$ 9.34} & \mbox{68.40 $\pm$ 8.08} & \mbox{67.11 $\pm$ 8.39} \\
        GPSL  & \mbox{\textbf{84.71} $\pm$ 0.29} & \mbox{\textbf{84.28} $\pm$ 0.20} & \mbox{\textbf{84.30} $\pm$ 0.32} & \mbox{\textbf{84.40} $\pm$ 0.51} \\
        \bottomrule
    \end{tabularx}

    \vspace{1.5em}

    \caption{Mild non-IID Test Accuracy $(C=5, \alpha=3.0)$}
    \label{tab:results_cifar10_mild}
    \begin{tabularx}{\columnwidth}{l*{4}{>{\centering\arraybackslash}X}}
        \toprule
        \textbf{Method} & $\mathbf{K=16}$ & $\mathbf{K=32}$ & $\mathbf{K=64}$ & $\mathbf{K=128}$ \\
        \midrule
        FLS   & \mbox{66.12 $\pm$ 6.92} & \mbox{70.97 $\pm$ 5.32}  & \mbox{70.45 $\pm$ 6.87} & \mbox{69.33 $\pm$ 10.62} \\
        FPLS  & \mbox{59.18 $\pm$ 14.48} & \mbox{64.61 $\pm$ 10.05} & \mbox{71.43 $\pm$ 4.92}  & \mbox{71.25 $\pm$ 6.08} \\
        GPSL  & \mbox{\textbf{84.50} $\pm$ 0.56} & \mbox{\textbf{84.34} $\pm$ 0.34} & \mbox{\textbf{84.52} $\pm$ 0.24} & \mbox{\textbf{84.43} $\pm$ 0.26} \\
        \bottomrule
    \end{tabularx}
    \vspace{1em}

    \parbox{\columnwidth}{\small%
    ResNet-18 on CIFAR-10: Top-1 test accuracy under IID and non-IID settings for different numbers of clients $K$. CL achieved $84.06\,\pm\,0.53$. \textbf{Bold} indicates the best per column.
    }
\end{table}

\begin{figure}[!t]
      \centering
      \includegraphics[width=\columnwidth]{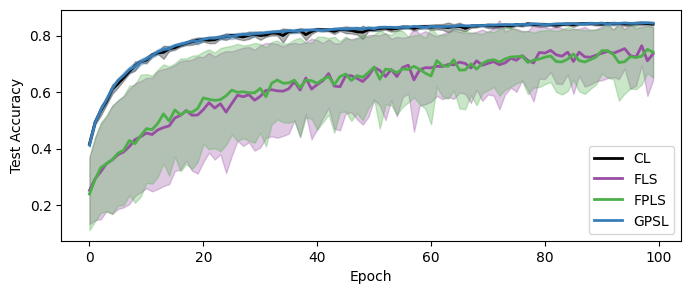}
      \caption{ResNet-18 on CIFAR-10: Test accuracy curves for each sampling method in a severe non-IID setting with standard deviation shaded.}
      \label{fig:accuracy_curve_cifar10}
\end{figure}

\Cref{tab:results_cifar10_severe} shows the performance of each sampling method under severe non-IID conditions. \Cref{tab:results_cifar10_mild} shows the performance in a mild non-IID scenario. In both cases, GPSL achieves a test accuracy on par with centralized learning, while FPLS and FLS degrade significantly. The results are consistent across both mild and severe non-IID scenarios for varying client counts $K$, indicating scalability and robustness to the degree of data heterogeneity. \Cref{fig:accuracy_curve_cifar10} shows the test accuracy curves for each sampling method under severe non-IID conditions. GPSL maintains a stable convergence curve, while FPLS and FLS exhibit significant fluctuations.

\renewcommand{\arraystretch}{1.2}
\begin{table}[t]
    \centering
    \caption{Effect of global batch size on test accuracy}
    \label{tab:cifar10_batch_size_severe}
    \begin{tabularx}{\columnwidth}{l*{3}{>{\centering\arraybackslash}X}}
        \toprule
        \textbf{Method} & $\mathbf{B=64}$ & $\mathbf{B=128}$ & $\mathbf{B=256}$ \\
        \midrule
        FLS   & \mbox{69.57 $\pm$ 2.93} & \mbox{67.73 $\pm$ 9.29} & \mbox{64.39 $\pm$ 4.80} \\
        FPLS  & \mbox{69.25 $\pm$ 4.96} & \mbox{67.11 $\pm$ 8.39} & \mbox{64.25 $\pm$ 11.19} \\
        GPSL  & \mbox{86.06 $\pm$ 0.28} & \mbox{\textbf{84.40} $\pm$ 0.51} & \mbox{\textbf{82.22} $\pm$ 0.45} \\
        CL    & \mbox{\textbf{86.29} $\pm$ 0.24} & \mbox{84.06 $\pm$ 0.53} & \mbox{81.47 $\pm$ 0.80} \\
        \bottomrule
    \end{tabularx}

    \vspace{1em}

    \parbox{\columnwidth}{\small%
    ResNet-18 on CIFAR-10: Top-1 test accuracy under severe non-IID setting for different global batch sizes $B$ with $K=128$ clients. \textbf{Bold} indicates the best per column.
    }
\end{table}

\subsection{Global Batch Size}
\label{sec:global_batch_size}
\Cref{tab:cifar10_batch_size_severe} shows performance results for the different sampling methods with varying global batch sizes $B$. GPSL consistently outperforms FPLS and FLS across all batch sizes, demonstrating its robustness to batch size variations. In particular, GPSL enables flexibility in choosing the global batch size whereas FPLS and FLS are sensitive to the choice of batch size and may result in much larger global batch sizes than intended due to rounding and the client data depletion problem (\Cref{sec:comparison_fixed_local_batches}).

\subsection{Batch Deviation}
\label{sec:batch_deviation}

\begin{figure}[!t]
      \centering
      \includegraphics[width=\columnwidth]{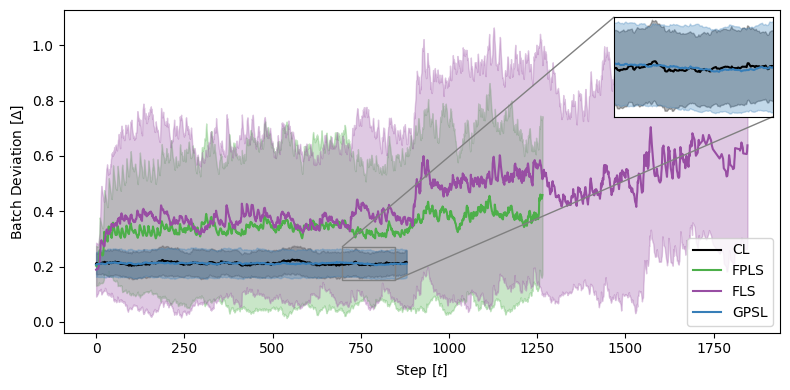}
      \caption{ResNet-18 on CIFAR-10: Batch deviation curves for each sampling method in a severe non-IID setting with exponential moving average smoothing and shaded standard deviation.}
      \label{fig:batch_deviation}
\end{figure}

\Cref{fig:batch_deviation} shows the empirical evaluation of the batch deviation per-step as defined in \cref{deviation_analysis} for each sampling method under severe non-IID conditions. GPSL maintains a low and stable batch deviation almost matching CL, while FPLS and FLS exhibit significant fluctuations, particularly in the standard deviation. This aligns with our theoretical analysis and performance results, indicating that the batch deviation is the main reason for the performance degradation of FPLS and FLS. 

Note that FLS and FPLS have more training steps than GPSL. This is due to the client data depletion problem, which leads to a larger number of batches per epoch with uneven batch sizes.

\subsection{Runtime}
\label{sec:runtime}

\begin{figure}[!t]
      \centering
      \includegraphics[width=\columnwidth]{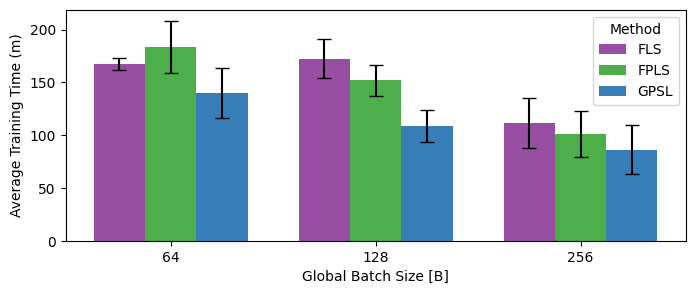}
      \caption{ResNet-18 on CIFAR-10: Average total training time for each sampling method in a severe non-IID setting, measured in minutes.}
      \label{fig:runtime}
\end{figure}

\Cref{fig:runtime} shows average training time for different global batch sizes $B$. GPSL consistently achieves a lower training time compared to FPLS and FLS. In particular when the global batch size is smaller. The computational overhead introduced by GPSL is thus negligible compared to the total training time saved. This result highlights that local sampling strategies can lead to a significant increase in the total training time due to the client data depletion problem. In particular, when the global batch size is smaller than the total number of clients.

\subsection{Limitations}
In settings with $K>B$, some clients necessarily receive zero allocations in a step, reducing instantaneous utilization and potentially inflating wall-clock time. Lightweight grouping or clustering of clients can mitigate this idle capacity \cite{liuWirelessDistributedLearning2023}. Our experiments assume a static, synchronous, straggler-free population, whereas real deployments must handle churn, delayed or dropped contributions, and network jitter. Because GPSL only changes how samples are allocated, established methods, such as asynchronous scheduling or deadline/slack participation remain directly compatible to address these systems issues \cite{chenAsynchronousOnlineFederated2020}. However, designing novel approaches for dealing with such instable situations specific to PSL remains an open research direction. Another direction for future work is to integrate churn-aware re-sampling and adaptive policies that jointly tune the global sampling and participation pattern in highly dynamic environments such as edge and IoT deployments.

\section{Conclusion} \label{conclusion}

We introduced parallel split learning with global sampling (GPSL), a drop‑in sampling mechanism that fixes the global batch size and allocates per‑client batch size schedules using pooled‑level proportions while clients sample locally. By eliminating per‑class rounding and decoupling the effective batch size from the number of participating clients, GPSL recovers the statistics of centralized uniform sampling without replacement and admits finite‑population deviation guarantees. Empirically, under non‑IID splits GPSL consistently attains centralized‑like accuracy and stabilizes optimization. Furthermore, GPSL shortens training time by avoiding inflation of steps induced by client data depletion without increasing server memory or coordination overhead. Relative to fixed local batching, GPSL changes only \/how batches are formed\/ while preserving the PSL training loop, thereby removing the additive bias that skews global label histograms and inflates the effective batch size. The main limitations are scope and setting: our evaluation assumes a static, synchronous population. When \(K>B\), per‑step idleness may limit utilization. These constraints are orthogonal to GPSL and can be addressed with existing PSL techniques. Looking ahead, we see value in churn‑aware re‑sampling and adaptive policies that jointly tune global batch size and participation. Collectively, these properties make GPSL practical for edge and IoT deployments, where resources are limited and client populations are large and heterogeneous

\section*{Acknowledgment}
This work was funded by the German Research Foundation
(DFG) under the Cluster of Excellence CARE: Climate-
Neutral And Resource-Efficient Construction (EXC 3115),
project number 533767731 and by the Federal Ministry of Re-
search, Technology, and Space (BMFTR, Germany) as part of
NeuroSys: Efficient AI-methods for neuromorphic computing
in practice (Projekt D) - under Grant 03ZU2106DA.

The authors used language-assistance tools for copy-editing; responsibility for the content rests with the authors.

\appendix

\subsection{Additional Results on CIFAR-100 and ResNet-34}
\label{app:additional_results}
We provide additional experiments on the CIFAR-100 dataset \cite{krizhevskyLearningMultipleLayers2009} and a deeper ResNet-34 backbone \cite{heDeepResidualLearning2016}. All experimental settings follow \Cref{sec:simulation_results} except for a longer training horizon of \(E{=}200\) epochs. Results are averaged over 3 runs and reported with standard deviation. 

\begin{figure}[!t]
      \centering
      \includegraphics[width=\columnwidth]{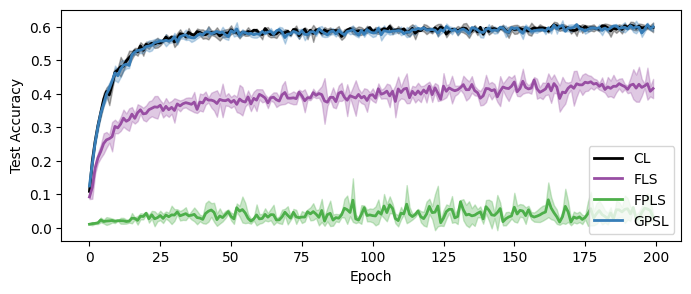}
  \caption{ResNet-34 on CIFAR-100: Test accuracy curves for each sampling method in a severe non-IID setting with standard deviation shaded.}
  \label{fig:acc_curve_res34_c100}
\end{figure}

\begin{figure}[!t]
      \centering
      \includegraphics[width=\columnwidth]{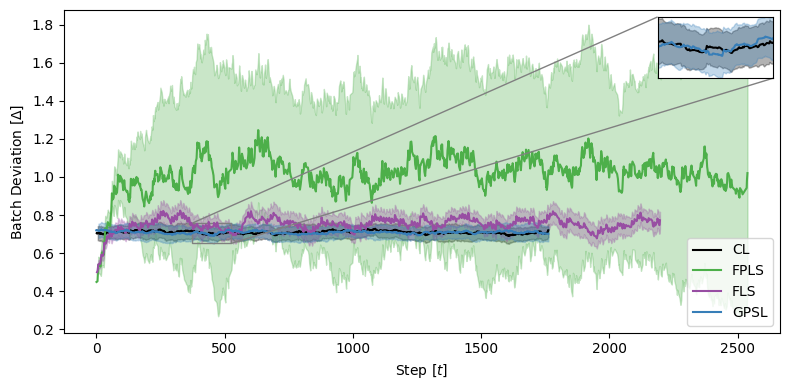}
  \caption{ResNet-34 on CIFAR-100: Batch deviation curves for each sampling method in a severe non-IID setting with exponential moving average smoothing and shaded standard deviation.}
  \label{fig:dev_curve_res34_c100}
\end{figure}

\Cref{tab:resnet34_accuracies} summarizes results in the severe non-IID case (\(C{=}2,\alpha{=}3.0\)) with \(K{=}64\) and \(B{=}128\). GPSL again matches CL on both datasets, while FLS and FPLS trail, with a larger gap on CIFAR-100. In \Cref{fig:acc_curve_res34_c100}, GPSL’s test accuracy on CIFAR-100 rises smoothly to a higher plateau with tight variability, whereas FLS and FPLS settle lower and fluctuate more. \Cref{fig:dev_curve_res34_c100} shows GPSL’s batch deviation closely tracking CL, while FLS and FPLS remain higher and more volatile. Notably, relative to the CIFAR-10/ResNet-18 experiment, FLS exhibits a lower batch deviation yet still shows a pronounced accuracy gap. This reflects increased task difficulty and model depth: small but persistent distributional mismatch spread across many classes over a longer training horizon can impede representation learning, so lower absolute deviation values need not translate into comparable accuracy. Collectively, these observations indicate GPSL’s suitability as a robust drop-in choice across architectures and datasets.

\renewcommand{\arraystretch}{1.2}
\begin{table}
    \centering
    \caption{Severe Non-IID Test Accuracy}
    \label{tab:resnet34_accuracies}
    \begin{tabularx}{\columnwidth}{l*{2}{>{\centering\arraybackslash}X}}
        \toprule
        \textbf{Method} & \textbf{CIFAR-10} & \textbf{CIFAR-100} \\
        \midrule
    FLS   & \mbox{25.42 $\pm$ 7.42} & \mbox{41.61 $\pm$ 2.88} \\
    FPLS  & \mbox{13.62 $\pm$ 2.75} & \mbox{3.13 $\pm$ 1.24} \\
    GPSL  & \mbox{\textbf{89.69} $\pm$ 0.24} & \mbox{\textbf{60.04} $\pm$ 1.18} \\
        \bottomrule
        \vspace{0.01em}
    \end{tabularx}
    \parbox{\columnwidth}{\small%
    ResNet-34 on CIFAR-10 and CIFAR-100: Top-1 test accuracy under a severe non-IID setting. CL achieved $ 89.62  \pm  1.16 $ on CIFAR-10 and $ 59.86  \pm  1.45 $ on CIFAR-100. \textbf{Bold} indicates the best per column.
    }
\end{table}

\bibliographystyle{IEEEtran}
\bibliography{IEEEabrv, references}

\end{document}